\pdfoutput=1

\documentclass[11pt]{article}
\usepackage{authblk}
\usepackage{naacl2021}

\usepackage{times}
\usepackage{latexsym}
\usepackage{amsmath,bm}
\usepackage{enumitem}
\usepackage{multirow}
\usepackage{multicol}
\usepackage{subcaption}
\usepackage{amssymb}
\usepackage{booktabs}
\usepackage{array}
\usepackage[pdftex]{graphicx}
\usepackage{url}
\usepackage{xcolor}
\usepackage{enumerate}
\usepackage{cleveref}
\usepackage{makecell}
\usepackage{textcomp}
\usepackage{float}

\crefformat{section}{\S#2#1#3}
\renewcommand{\footnotesize}{\fontsize{8.5pt}{10.5pt}\selectfont}
\newcommand{\floor}[1]{\lfloor #1 \rfloor}

\usepackage[T1]{fontenc}

\usepackage[utf8]{inputenc}

\usepackage{microtype}

\title{Distantly Supervised Relation Extraction with \\ Sentence Reconstruction and Knowledge Base Priors}

\author[1]{Fenia Christopoulou}
\author[2,3]{Makoto Miwa}
\author[1]{Sophia Ananiadou}
\affil[1]{National Centre for Text Mining, \authorcr
\textnormal{\normalsize Department of Computer Science, The University of Manchester, United Kingdom}}
\affil[2]{Toyota Technological Institute, Nagoya, 468-8511, Japan}
\affil[3]{Artificial Intelligence Research Center, National Institute of Advanced Industrial Science and Technology, Japan}
\affil[ ]{\tt \small \{efstathia.christopoulou, sophia.ananiadou\}@manchester.ac.uk}
\affil[ ]{\tt \small makoto-miwa@toyota-ti.ac.jp}

\begin{document}
\maketitle
\begin{abstract}
We propose a multi-task, probabilistic approach to facilitate distantly supervised relation extraction by bringing closer the representations of sentences that contain the same Knowledge Base pairs. To achieve this, we bias the latent space of sentences via a Variational Autoencoder (\textsc{vae}) that is trained jointly with a relation classifier. The latent code guides the pair representations and influences sentence reconstruction.
Experimental results on two datasets created via distant supervision indicate that multi-task learning results in performance benefits. Additional exploration of employing Knowledge Base priors into the \textsc{vae} reveals that the sentence space can be shifted towards that of the Knowledge Base, offering interpretability and further improving results\footnote{Source code is available at \url{https://github.com/fenchri/dsre-vae}}.
\end{abstract}

\section{Introduction}

\begin{figure}[t!]
    \centering
    \includegraphics[width=\linewidth]{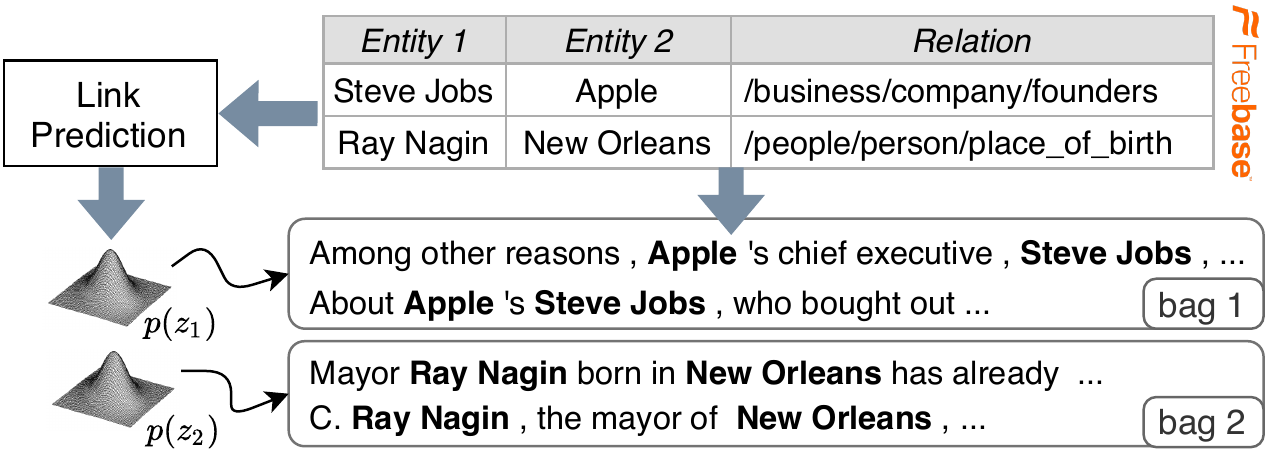}
    \caption{Example of the bag-level setting in distantly supervised relation extraction and the main idea of our approach. Sentences are adapted from the \textsc{nyt10} dataset~\citep{riedel2010modeling}.}
    \label{fig:example}
\end{figure}

Distant supervision (DS) is a setting where information from existing, structured knowledge, such as Knowledge Bases (KB), is exploited to automatically annotate raw data.
For the task of relation extraction, this setting was popularised by \citet{mintz2009distant}. Sentences containing a pair of interest were annotated as positive instances of a relation, if and only if the pair was found to share this relation in the KB.
However, due to the strictness of this assumption, relaxations were proposed, such as the at-least-one assumption introduced by \citet{riedel2010modeling}: Instead of assuming that all sentences in which a known related pair appears express the relationship, we assume that at least one of these sentences (namely a \textit{bag} of sentences) expresses the relationship.
Figure \ref{fig:example} shows example bags for two entity pairs.

The usefulness of distantly supervised relation extraction (DSRE) is reflected in facilitating automatic data annotation, as well as the usage of such data to train models for KB population~\citep{ji2011knowledge}. However, DSRE suffers from noisy instances, long-tail relations and unbalanced bag sizes. 
Typical noise reduction methods have focused on using attention~\citep{lin2016neural,ye2019intra-inter} or reinforcement learning~\citep{qin2018robust,wu2019improving}. For long-tail relations, relation type hierarchies and entity descriptors have been proposed~\citep{she2018descriptions,zhang2019long-tail,hu2019improving}, while the limited bag size is usually tackled through incorporation of external data~\citep{beltagy2019distant-direct}, information from KBs~\citep{vashishth2018reside} or pre-trained language models~\citep{alt2019fine}. 
Our goal is not to investigate noise reduction, since it has already been widely addressed. Instead, we aim to propose a more general framework that can be easily combined with existing noise reduction methods or pre-trained language models.

Methods that combine information from Knowledge Bases in the form of pre-trained Knowledge Graph (KG) embeddings have been particularly effective in DSRE. This is expected since they capture broad associations between entities, thus assisting the detection of facts. Existing approaches either encourage explicit agreement between sentence- and KB-level classification decisions~\citep{weston2013connecting,xu2019hrere}, minimise the distance between KB pairs and sentence embeddings~\citep{wang2018label} or directly incorporate KB embeddings into the training process in the form of attention queries~\citep{han2018jointnre,she2018descriptions,hu2019improving}. Although these signals are beneficial, direct usage of KB embeddings into the model often requires explicit KB representations of entities and relations, leading to poor generalisation to unseen examples. In addition, forcing decisions between KB and text to be the same makes the connection between context-agnostic (from the KB) and context-aware (from sentences) pairs rigid, as they often express different things.

Variational Autoencoders (\textsc{vae}s)~\citep{kingma2013auto} are latent variable encoder-decoder models that parameterise posterior distributions using neural networks. As such, they learn an effective latent space which can be easily manipulated. 
Sentence reconstruction via encoder-decoder networks helps sentence expressivity by learning semantic or syntactic similarities in the sentence space. On the other hand, signals from a KB can assist detection of factual relations. 
We aim to combine these two using a \textsc{vae} together with a bag-level relation classifier. We then either force each sentence's latent code to be close to the Normal distribution~\citep{bowman2016generating}, or to a prior distribution obtained from KB embeddings. This latent code is employed into sentence representations for classification and is responsible for sentence reconstruction. As it is influenced by the prior we essentially inject signals from the KB to the target task.
In addition, sentence reconstruction learns to preserve elements that are useful for the bag relation.
To the best of our knowledge, this is the first attempt to combine a \textsc{vae} with a bag-level classifier for DSRE.

Finally, there are methods for DSRE that follow a rather flawed evaluation setting, where several test pairs are included in the training set. Under this setting, the generalisability of such methods can be exaggerated.
We test these approaches under data without overlaps and find that their performance is severely deprecated. With this comparison, we aim to promote evaluation on the amended version of existing DSRE data that can prevent memorisation of test pair relations. 
Our contributions are threefold: \vspace{-0.23cm}
\begin{itemize}[noitemsep,leftmargin=*]
    \item Propose a multi-task learning setting for DSRE. Our results suggest that combination of both bag classification and bag reconstruction improves the target task.
    
    \item Propose a probabilistic model to make the space of sentence representations resemble that of a KB, promoting interpretability.
    
    \item Compare existing approaches on data without train-test pair overlaps to enforce fairer comparison between models.
\end{itemize}

\section{Proposed Approach}

\subsection{Task Description}
In DSRE, the bag setting is typically adopted. A model's input is a pair of named entities $e_1, e_2$ (mapped to a Knowledge Base), and a bag of sentences $B = \{s_1, s_2, \dots, s_n \}$, where the pair occurs, retrieved from a raw corpus. The goal of the task is to identify the relation(s), from a pre-defined set $R$, that the two entities share, based on the sentences in the bag $B$. Since each pair can share multiple relations at the same time, the task is considered a multi-label classification problem.

\subsection{Overall Framework}
\begin{figure}[t!]
    \centering
    \includegraphics[width=0.75\linewidth]{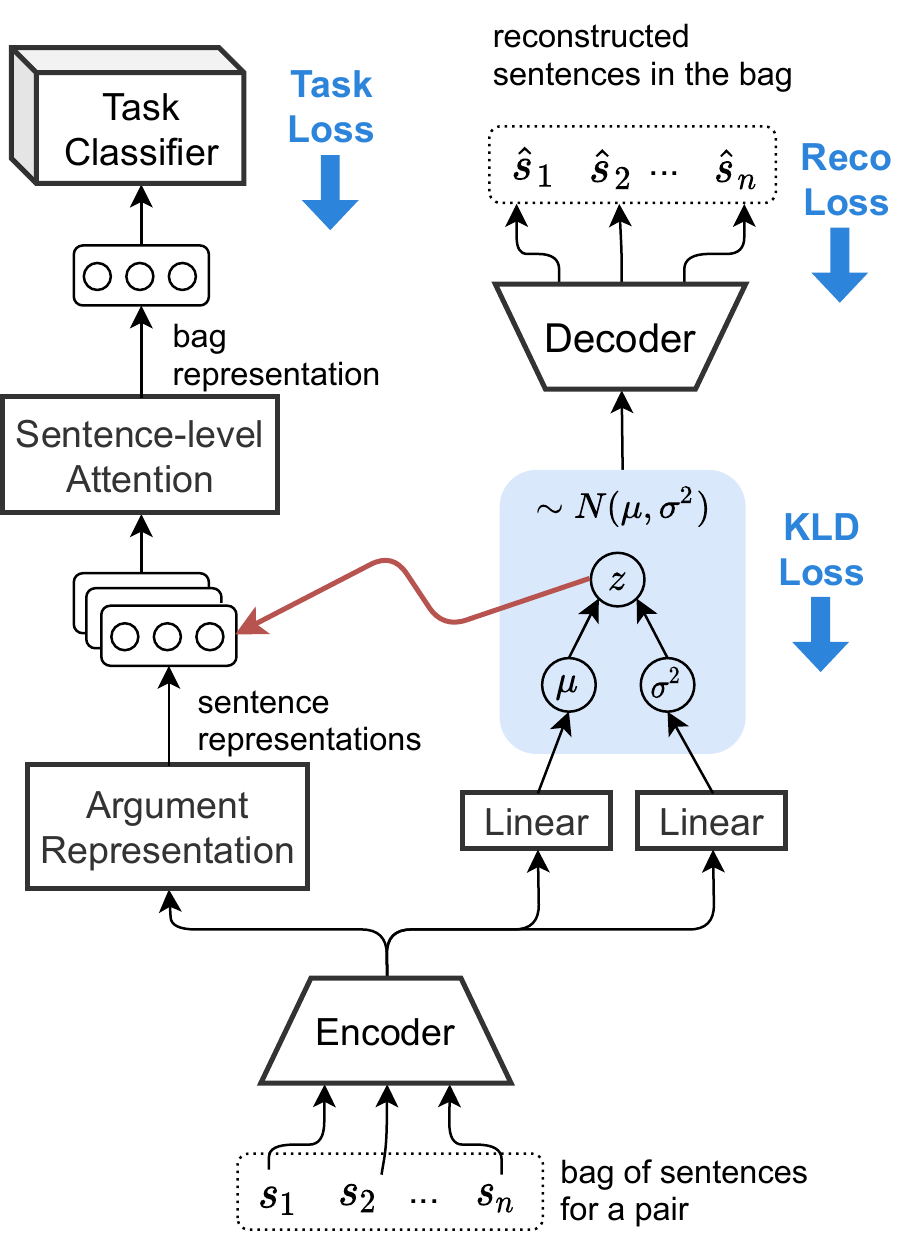}
    \caption{Schematic of the model architecture.}
    \label{fig:arch}
\end{figure}

Our proposed approach is illustrated in Figure~\ref{fig:arch}. 
The main goal is to create a joint learning setting where a bag of sentences is encoded and reconstructed and, at the same time, the bag representation is used to predict relation(s) shared between two given entities.
The architecture receives as input a bag of sentences for a given pair and outputs (i) predicted relations for the pair and (ii) the reconstructed sentences in the bag.
The two outputs are produced by two branches: the left branch, corresponding to bag classification and the right branch, corresponding to bag reconstruction. 
Both branches start from a shared encoder and they communicate via the latent code of a \textsc{vae} that is responsible for the information used in the representation and reconstruction of each sentence in the bag. Naturally, both branches have an effect on one another during training.


\subsection{Bag Reconstruction}
\label{sec:bag_reco}
Autoencoders~\citep{rumelhart1986autoencoders} are encoder-decoder neural networks that are trained in an unsupervised manner, i.e., to reconstruct their input (e.g. a sentence). They learn an informative representation of the input into a dense and smaller feature vector, namely the latent code. This intermediate representation is then used to fully reconstruct the original input.
Variational Autoencoders (\textsc{vae})~\citep{kingma2013auto} offer better generalisation capabilities compared to the former by sampling the features of the latent code from a prior distribution that we assume to be similar to the distribution of the data.

\subsubsection{Encoder}
\label{sec:encoder}
We form the input of the network similarly to previous work. Each sentence in the input bag is transformed into a sequence of vectors. Words and positions are mapped into real-valued vectors via word embedding $\mathbf{E}^{(w)}$ and position embedding layers $\mathbf{E}^{(p)}$, similarly to \citet{lin2016neural}. 
The concatenation of word 
($\mathbf{w}$) and position ($\mathbf{p}$) embeddings $\mathbf{x}_t = [\mathbf{w}_t ; \mathbf{p}_t^{(e_1)}; \mathbf{p}_t^{(e_2)}]$ 
forms the representation of each word in the input sentence. 
A Bidirectional Long-Short Term Memory (BiLSTM) network~\citep{hochreiter1997long} acts as the encoder, producing contextualised representations for each word.

The representations of the left-to-right and right-to-left passes of the BiLSTM are summed to produce the output representation of each word $t$, 
$\mathbf{o}_t = \overrightarrow{\mathbf{o}_t} + \overleftarrow{\mathbf{o}_t}$, 
as well as the representations of the last hidden $\mathbf{h} = \overrightarrow{\mathbf{h}} + \overleftarrow{\mathbf{h}}$ and cell states $\mathbf{c} = \overrightarrow{\mathbf{c}} + \overleftarrow{\mathbf{c}}$ of the input sentence. 
We use the last hidden and cell states of each sentence $s$ to construct the parameters of a posterior distribution $q_\phi(\mathbf{z}|\mathbf{s})$ using two linear layers,
\begin{equation}
\begin{aligned}
    \bm{\mu} &= \mathbf{W}_{\mu} [\mathbf{h}; \mathbf{c}] + \mathbf{b}_{\mu}, \\ 
    \bm{\sigma}^2 &= \mathbf{W}_{\sigma} [\mathbf{h}; \mathbf{c}] + \mathbf{b}_{\sigma}, \\
\end{aligned}
\label{eq:mu_sigma}
\end{equation}
where $\bm{\mu}$ and $\bm{\sigma}^2$ are the parameters of a multivariate Gaussian, representing the feature space of the sentence.
This distribution is approximated via a latent code $\mathbf{z}$, using the reparameterisation trick~\citep{kingma2013auto} to enable back-propagation, as follows:
\begin{equation}
\mathbf{z} = \bm{\mu} + \bm{\sigma} \odot \bm{\epsilon}, \; \text{where} \; \bm{\epsilon} \sim \mathcal{N}(\bm{0}, \mathbf{I}).
\label{eq:z}
\end{equation}
This trick essentially forms the posterior as a function of the normal distribution.

\subsubsection{Decoder}
\label{sec:decoder}
The decoder network is a uni-directional LSTM network, that reconstructs each sentence in the input bag.
The input is formed in two steps.
Firstly, the latent code $\mathbf{z}$ is given as the initial hidden state of the decoder $\mathbf{h}'_0$ via a linear layer transformation.
Secondly, the same latent code is concatenated with the representation of each word $\mathbf{w}_t$ in the input sequence of the decoder. 
\begin{equation}
    \mathbf{h}'_0 = \mathbf{W} \mathbf{z} + \mathbf{b}, \; \;
    \mathbf{x}'_t = [ \mathbf{w}_t; \mathbf{z} ],
\end{equation}
A percentage of words in the decoder's input is randomly replaced by the UNK word to force the decoder to rely on the latent code for word prediction, similar to \citet{bowman2016generating}.

\subsubsection{Learning}
\label{sec:learning}
The optimisation objective of the \textsc{vae}, namely Evidence Lower BOund (ELBO), is the combination of two losses. The first is the reconstruction loss that corresponds to the cross entropy between the actual sentence $s$ and its reconstruction $\hat{s}$.
The second is the Kullback-Leibler divergence ($D_\text{KL}$) between a prior distribution $p_\theta(\mathbf{z})$, which the latent code is assumed to follow, and the posterior $q_\phi(\mathbf{z}|\mathbf{h})$, which the decoder produces,
\begin{multline}
    L_\text{ELBO} = \mathbb{E}_{z \sim q_\phi(z|h)} \left[ \log(p_\theta(\mathbf{h}|\mathbf{z})) \right] \\
    - D_\text{KL}\left( q_\phi(\mathbf{z}|\mathbf{h}) || p_\theta(\mathbf{z}) \right)
\end{multline}
The first loss is responsible for the accurate reconstruction of each word in the input, while the second acts as a regularisation term that encourages the posterior of each sentence to be close to the prior.
Typically, an additional parameter $\beta$ is introduced in front of the $D_\text{KL}$ to overcome KL vanishing, a phenomenon where the posterior collapses to the prior and the \textsc{vae} essentially behaves as a standard autoencoder~\citep{bowman2016generating}.

\subsection{Bag Classification}

Moving on to the left branch of Figure \ref{fig:arch}, in order to represent a bag we first need to represent each sentence inside it.
We realise this using information produced by the \textsc{vae} as follows.

\subsubsection{Sentence Representation}
Given the contextualised output of the encoder $\mathbf{o}$, we construct entity representations $\mathbf{e}_1$ and $\mathbf{e}_2$ for a given pair in a sentence by averaging the word representations included in each entity.
A sentence representation $\mathbf{s}$ is formed as follows: 
\begin{equation}
\begin{aligned}
    \mathbf{e}_i = \frac{1}{|e_i|} \sum_{k \in e_i} \mathbf{o}_k, \; \;
    \mathbf{s} = \mathbf{W}_v [\mathbf{z}; \mathbf{e}_1; \mathbf{e}_2], 
\end{aligned}
\end{equation}
where $|e_i|$ corresponds to the number of words inside the mention span of entity $e_i$ and $\mathbf{z}$ is the latent code of the sentence that was produced by the \textsc{vae}, as described in Equation (\ref{eq:z}).

\subsubsection{Bag Representation}
\label{sec:bag_enc}
In order to form a unified bag representation $B$ for a pair, we adopt the popular selective attention approach introduced by \citet{lin2016neural}. 
In particular, we first map relations into real-valued vectors, via a relation embedding layer $\mathbf{E}^{(r)}$. 
Each relation embedding is then used as a query over the sentences in the bag, resulting in $|R|$ bag representations for each pair,  
\begin{equation}
    a_r^{(s_i)} = \frac{\exp{ (\mathbf{s}_i^\top \mathbf{r})}}{\sum\limits_{j \in B} \exp{(\mathbf{s}_j^\top \mathbf{r})}}, \ \
    \mathbf{B}_r = \sum_{i=1}^{|B|} a_r^{(s_i)} \mathbf{s}_i,
\end{equation}
where $\mathbf{r}$ is the embedding associated with relation $r$, $\mathbf{s}_i$ is the representation of sentence $s_i \in B$, $a_r^{(s_i)}$ is the weight of sentence $s_i$ with relation $r$ and $\mathbf{B}_r$ is the final bag representation for relation $r$.

During classification, we select the probability of predicting a relation category $r$, using the bag representation that was constructed when the respective relation embedding $\mathbf{r}$ was the query. 
Binary cross entropy loss is applied on the resulting predictions,
\begin{equation}
\begin{aligned}
    p(r = 1|B) &= \sigma( \mathbf{W}_c \; \mathbf{B}_r + \mathbf{b}_c), \\ 
    L_\text{BCE} &= - \sum_{r} y_r \log p(r|B) \\
    & + (1 - y_r) \log (1 - p(r|B)),
\end{aligned}
\end{equation}
where $\mathbf{W}_c$ and $\mathbf{b}_c$ are learned parameters of the classifier, $\sigma$ is the sigmoid activation function, $p(r|B)$ is the probability associated with relation $r$ given a bag $B$ and $y_r$ is the ground truth for this relation with possible values 1 or 0.

\subsection{Knowledge Base Priors}
In the scenario where no KB information is incorporated into the model, we simply assume that the prior distribution of the latent code $p_\theta(\mathbf{z})$ is a standard Gaussian with zero mean and identity covariance $\mathcal{N}(\bm{0}, \mathbf{I})$.

To integrate information about the nature of triples into the bag-level classifier, we create KB-guided priors as an alternative to the standard Gaussian. 
In particular, we train a link prediction model, such as TransE~\citep{bordes2013translating}, on a subset of the Knowledge Graph that was used to originally create the dataset. 
Using the link prediction model, we obtain entity embeddings for the subset KB.
A KB-guided prior can thus be constructed for each pair, as another Gaussian distribution with mean value equal to the KB pair representation and covariance as the identity matrix,
\begin{equation}
    p_\theta(\mathbf{z}) \sim \mathcal{N}(\bm{\mu}_\textsc{kb}, \mathbf{I}), \; \text{with} \; \; \bm{\mu}_\textsc{kb} = \mathbf{e}_h - \mathbf{e}_t,
    \label{eq:mu_kb}
\end{equation}
where $\mathbf{e}_h$ and $\mathbf{e}_t$ are the vectors for entities $e_\text{head}$ and $e_\text{tail}$ as resulted from training a link prediction algorithm on a KB.

The link prediction algorithm is trained to make representations of pairs expressing the same relations to be close in space. 
Hence, by using KB priors we try to force the distribution of sentences in a bag to follow the distribution of the pair in the KB.
If one of the pair entities does not exist in the KB subset, the mean vector of the pair's prior will be zero, resulting in a standard Gaussian prior.
Finally, KB priors are only used during training. Consequently, the model does not use any direct KB information during inference.

\subsection{Training Objective}
We train jointly bag classification and sentence reconstruction. 
The final optimisation objective is formed as,
\begin{equation}
L = \lambda \; L_\text{BCE} + (1 - \lambda) L_\text{ELBO},
\end{equation}
where $\lambda$ corresponds to a weight in $[0, 1]$. We weigh the classification loss more than the ELBO to allow the model to better fit the target task.

\section{Experimental Settings}

\subsection{Datasets}
We experiment with the following two datasets:

\noindent \textbf{\textsc{nyt10}.} The widely used New York Times dataset~\citep{riedel2010modeling} contains $53$ relation categories including a negative relation (NA) indicating no relation between two entities.
We use the version of the data provided by the OpenNRE framework~\citep{han2019opennre}, which removes overlapping pairs between train and test data.
The dataset statistics are shown in Table \ref{tab:data_stats}. Additional information can be found in Appendix \ref{app:info_nyt}.

For the choice of the Knowledge Base, we use a subset of Freebase\footnote{\url{https://developers.google.com/freebase}} that includes $3$ million entities with the most connections, similar to \citet{xu2019hrere}. For all pairs appearing in the test set of \textsc{nyt10} (both positive and negative), we remove all links in the subset of Freebase to ensure that we will not memorise any relations between them~\citep{weston2013connecting}. The resulting KB contains approximately $24$ million triples. \\

\noindent \textbf{\textsc{WikiDistant}.} The WikiDistant dataset is almost double the size of the \textsc{nyt10} and contains $454$ target relation categories, including the negative relation. It was recently introduced by \citet{han2020more} as a cleaner and more well structured bag-level dataset compared to \textsc{nyt10}, with fewer negative instances.

For the Knowledge Base, we use the version of Wikidata\footnote{\url{https://www.wikidata.org/}} provided by \citet{wang2019kepler} (in particular the transductive split\footnote{\url{https://deepgraphlearning.github.io/project/wikidata5m}}), containing approximately $5$ million entities. Similarly to Freebase, we remove all links between pairs in the test set from the resulting KB, which contains approximately $20$ million triples after pruning.

\begin{table}[t!]
    \centering
    \scalebox{0.8}{
    \begin{tabular}{clrrr}
    \toprule
        Dataset & Split & Instances & Bags & NA (\%)  \\ \midrule
        \multirow{3}{*}{\parbox[t]{2.5cm}{\textsc{nyt10} \\ \# Relations: 53}}
        & Train & 469,290  & 252,044 & 93.4 \\
        & Val.  & 53,321   & 28,109  & 93.5 \\
        & Test  & 172,448  & 96,678  & 97.9 \\ \midrule
                                        
        \multirow{3}{*}{\parbox[t]{2.5cm}{\textsc{WikiDistant} \# Relations: 454}}
        & Train & 1,050,246  & 575,620 & 64.8  \\
        & Val.  & 29,145     & 14,748  & 70.6 \\ 
        & Test  & 28,897     & 15,509  & 72.0 \\
    \bottomrule 
    \end{tabular}
    }
    \caption{Datasets statistics. `NA' correponds to the `no relation' category.}
    \label{tab:data_stats}
\end{table}

\subsection{Evaluation Metrics}
Following prior work, we consider the Precision-Recall Area Under the Curve (AUC) as the primary metric for both datasets. 
We additionally report Precision at $N$ (P@N), that measures the percentage of correct classifications for the top $N$ most confident predictions.

\subsection{Training}
To obtain the KB priors, we train TransE on the subsets of Freebase and Wikidata using the implementation of the DGL-KE toolkit~\citep{DGL-KE} for 500K steps and a dimensionality equal to the dimension of the latent code.
The main model was implemented with PyTorch~\citep{pytorch}.
We use the Adam~\citep{kingma2014adam} optimiser with learning rate $0.001$. KL logistic annealing is incorporated only in the case where the prior is the Normal distribution to avoid KL vanishing~\citep{bowman2016generating}. Early stopping is used to determine the best epoch based on the AUC score on the validation set.
Words in the vocabulary are initialised with pre-trained, $50$-dimensional GloVe embeddings~\citep{pennington2014glove}.

We limit the vocabulary size to the top 40K and 50K most frequent words for \textsc{nyt10} and \textsc{WikiDistant}, respectively. To enable fast training, we use Adaptive Softmax~\citep{grave2017adaptive}. The maximum sentence length is restricted to 50 for \textsc{nyt10} and 30 words for \textsc{WikiDistant}.
Each bag in the training set is allowed to contain maximum 500 sentences selected randomly. For prediction on the validation and test sets, all sentences (with full length) are used.

\begin{table*}[t!]
\centering
\scalebox{0.8}{
\begin{tabular}{llcccccccc}
    \toprule
    \multirow{4}{*}{Method} & \multirow{4}{*}{Encoder}
    & \multicolumn{4}{c}{NYT 520K} & \multicolumn{4}{c}{NYT 570K} \\ \cmidrule(lr){3-6} \cmidrule(lr){7-10}
    
    & & \multirow{2}{*}{AUC ({\small \%})} & \multicolumn{3}{c}{P@N ({\small \%})}
    & \multirow{2}{*}{AUC ({\small \%})} & \multicolumn{3}{c}{P@N ({\small \%})} \\
    \cmidrule(lr){4-6} \cmidrule(lr){8-10}
    
    & & & 100 & 200 & 300 & & 100 & 200 & 300 \\ \midrule
    Baseline 
    & \multirow{3}{*}{\small{BiLSTM}}	
    &  34.94
    & 74.0 &	67.5 &	67.0
    & 43.59
    & 84.0 &	77.0 &	75.3 \\
    
    \ \ $+$ $p_\theta(z) \sim \mathcal{N}(0, I)$	
    &
    & 38.59
    & 74.0 &	74.5 &	71.6
    &  44.64
    & 80.0 &	76.0 &	75.6 \\
     
    \ \ $+$ $p_\theta(z) \sim \mathcal{N}(\mu_\textsc{kb}, I)$
    & 
    & 42.89
    &  83.0 &	75.5 &	73.0
    & 45.52  
    & 81.0 & 77.5 & 73.6  \\   \midrule
    
    \textsc{pcnn-att}~\citep{lin2016neural} 
    & \small{PCNN}  
    & 32.66
    & 71.0	& 67.5	& 62.6 
    & 36.25
    & 76.0 &	72.5 &	64.0 \\
                                            
    \textsc{joint nre}~\citep{han2018jointnre} 
    & \small{CNN} 
    & 30.62	
    & 60.0 &	57.0 &	55.3 
    & 40.15
    & 75.8 & - & 68.0 \\
   
    \textsc{reside}~\citep{vashishth2018reside} 
    & \small{BiGRU} 
    & 35.80
    & 80.0 &	69.0 &	65.3
    & 41.60
    & 84.0 & 78.5 & 75.6 \\
                                                
    \textsc{intra-inter bag}~\citep{ye2019intra-inter} 
    & \small{PCNN} 
    & 	34.41
    &  82.0 &	74.0 &	69.0 
    & 42.20
    & 91.8 & 84.0 & 78.7 \\
    
    \textsc{distre}~\citep{alt2019fine} 
    & \small{GPT-2}
    & 42.20
    & 68.0	& 67.0 & 65.3 
    & - 
    & - & - & - \\
    
    \bottomrule
\end{tabular}
}
\caption{Performance comparison between different methods on the \textsc{nyt10} test set for the two different versions of the dataset. Results in the 520K column are re-runs of existing implementations, except for \textsc{distre}. Results on the 570K column are taken from the respective publications.}
\label{tab:nyt_res}
\end{table*}

\begin{table}[h!]
    \centering
    \scalebox{0.75}{
    \begin{tabular}{lcccc}
    \toprule
        \multirow{2}{*}{Method} & \multirow{2}{*}{AUC ({\small \%})}
        & \multicolumn{3}{c}{P@N ({\small \%})} \\ \cmidrule{3-5}
        & & 100 & 200 & 300 \\ \midrule
        
        Baseline 
        & 28.54 & 94.0 & 93.0 & 88.3 \\
        
        \ \ $+$ $p_\theta(z) \sim \mathcal{N}(0, I)$ 
        & 30.59 & 96.0 & 93.5 & 89.3 \\
        
        \ \ $+$ $p_\theta(z) \sim \mathcal{N}(\mu_\textsc{kb}, I)$ 
        & 29.54 & 92.0 & 89.0 & 90.0 \\ \midrule
        
        \textsc{pcnn-att}~\citep{han2020more}  
        & 22.20 & - & - & - \\ \midrule \midrule
        
        \multicolumn{5}{c}{\textit{w/o non KB-prior pairs (72\% of training pairs preserved)}} \\ \midrule
        
        Baseline 
        & 26.16 & 88.0 & 85.0 & 82.6 \\
        
        \ \ $+$ $p_\theta(z) \sim \mathcal{N}(0, I)$ 
        & 27.46 & 90.0 & 88.0 & 84.6 \\
        
        \ \ $+$ $p_\theta(z) \sim \mathcal{N}(\mu_\textsc{kb}, I)$ 
        & 28.38 & 94.0 & 95.0 & 89.3 \\ 
        \bottomrule
    \end{tabular}
    }
    \caption{Performance comparison on the \textsc{WikiDistant} test set.}
    \label{tab:wiki_res}
\end{table}

\subsection{Baselines}
In this work we compare with various models applied on the \textsc{nyt10} dataset:
\textsc{\textbf{pcnn-att}}~\citep{lin2016neural} is one of the first neural models that uses a PCNN encoder and selective attention over the instances in a bag, similar to our approach. 
\textsc{\textbf{reside}}~\citep{vashishth2018reside}, utilises syntactic, entity and relation type information as additional input to the network to assist classification. 
\textsc{\textbf{joint nre}}~\citep{han2018jointnre} jointly trains a textual relation extraction component and a link prediction component by sharing attention query vectors among the two. 
\textsc{\textbf{intra-inter bag}}~\citep{ye2019intra-inter} applies two attention mechanisms inside and across bags to enforce similarity between bags that share the same relations. 
\textbf{\textsc{distre}}~\citep{alt2019fine} uses a pre-trained Transformer model, instead of a recurrent or convolutional encoder, fine-tuned on the \textsc{nyt10} dataset. 

We report results on both the filtered data (520K) that do not contain train-test pair overlaps, as well as the non-filtered version (570K) to better compare with prior work\footnote{More information about the two versions can be found in Appendix \ref{app:info_nyt}}.
With the exception of \textsc{distre}, all prior approaches were originally applied on the 570K version. 
Hence, performance of prior work on the 520K version corresponds to re-runs of existing implementations (via their open-source code). For the non-filtered version, results are taken from the respective publications\footnote{For \textsc{pcnn-att} we re-run both the 520K and the 570K versions using the OpenNRE toolkit.}. 

For the \textsc{WikiDistant} dataset, we compare with the \textsc{\textbf{pcnn-att}} model as this is the only model currently applied on this data~\citep{han2020more}.
We also compare our proposed approach with two additional baselines. The first baseline model (Baseline) does not use the \textsc{vae} component at all. 
In this case the sentence representation is simply created using the last hidden state of the encoder, $\mathbf{s} = [\mathbf{h}; \mathbf{e}_1; \mathbf{e}_2]$, instead of the latent code.
The second model ($p_\theta(z) \sim \mathcal{N}(0, I)$) incorporates reconstruction with a standard Gaussian prior and the final model ($p_\theta(z) \sim \mathcal{N}(\mu_\textsc{kb}, I)$) corresponds to our proposed model with KB priors.

\section{Results}

\begin{figure}[t!]
    \centering
    \includegraphics[width=0.93\linewidth]{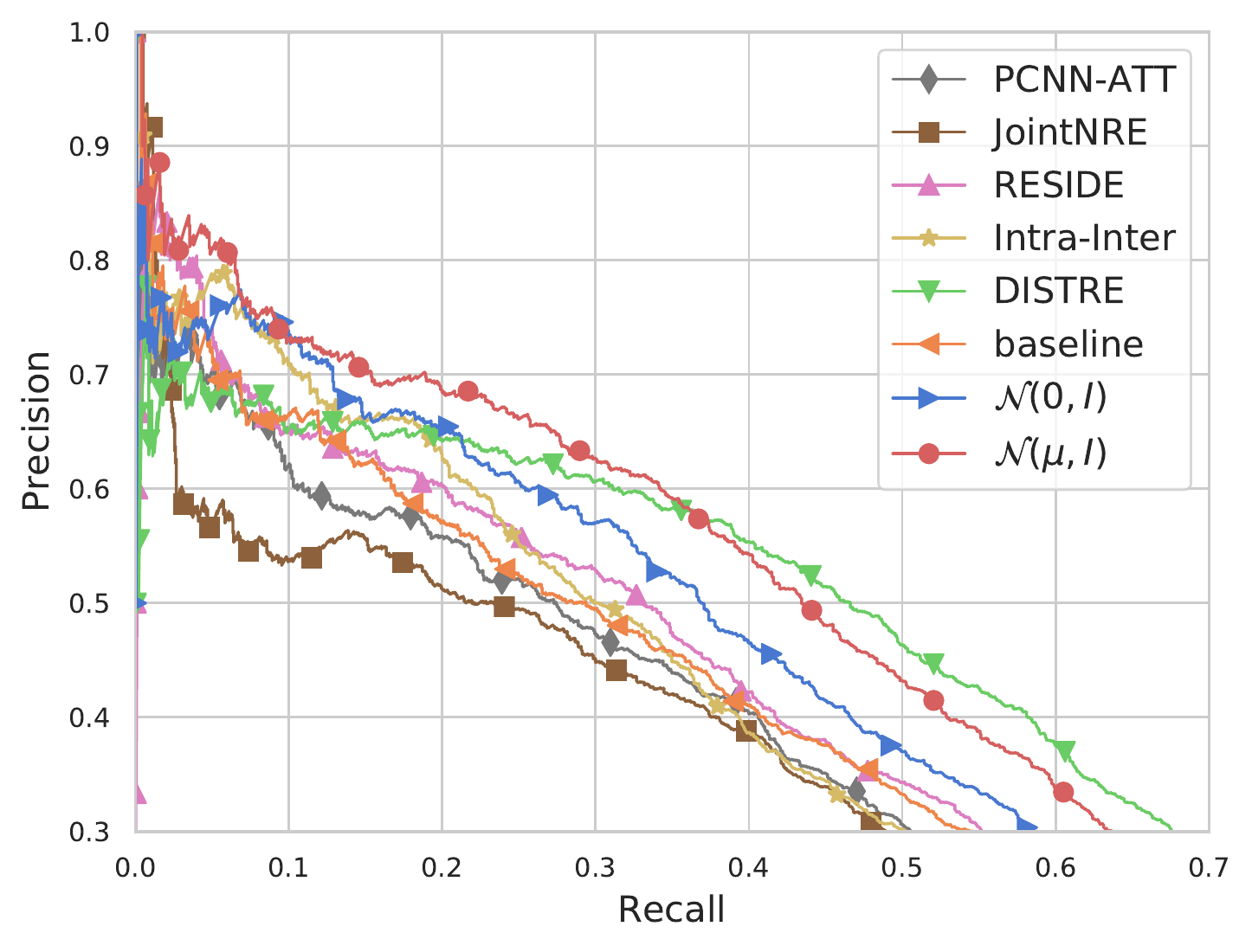}
    \caption{Precision-Recall curves for the \textsc{nyt10} (520K version) test set.}
    \label{fig:pr_nyt_520}
\end{figure}

The results of the proposed approach versus existing methods on the \textsc{nyt10} dataset are shown in Table \ref{tab:nyt_res}.
The addition of reconstruction further improves performance by 3.6 percentage points (pp), while KB priors offer an additional of 4.3pp.
Compared with \textsc{distre}, our model achieves comparable performance, even if it does not use a pre-trained language model. As we observe from the precision-recall curve in Figure \ref{fig:pr_nyt_520}, our model is competitive with \textsc{distre} for up to 35\% of the recall range but for the tail of the distribution a pre-trained language model has better results. This can be attributed to the world knowledge it has obtained via pre-training, which is much more vast than a KB subset. 
Overall, for the reduced version of the dataset \textsc{vae} with KB-guided priors surpasses the entire recall range of all previous methods.
For the 570K version, our model is superior to other approaches in terms of AUC score, even for the baseline. We speculate this is because we incorporate argument representations into the bag representation. As a result, overlapping pairs between training and test set have learnt strong argument representations.

Regarding the results on the \textsc{WikiDistant} dataset in Table \ref{tab:wiki_res}, once again we observe that reconstruction helps improve performance. However, it appears that KB priors have a negative effect. 
We find that in the \textsc{nyt10} dataset 96\% of the training pairs are associated with a prior. Instead, this portion is only 72\% for \textsc{wikidistant}. The reason for this discrepancy could be the reduced coverage that potentially causes a confusion between the two signals\footnote{If a pair does not have a KB prior it will be assigned the Normal prior instead.}.
To test this hypothesis, we re-run our models on a subset of the training data, removing pairs that do not have a KB prior. As observed in the second half of Table~\ref{tab:wiki_res}, priors do seem to have a positive impact under this setting, indicating the importance of high coverage in prior-associated pairs. We use this setting for the remainder of the paper.

\section{Analysis}
\begin{figure*}[t!]
    \centering
    \begin{subfigure}[b]{0.25\textwidth}
    \includegraphics[width=\linewidth]{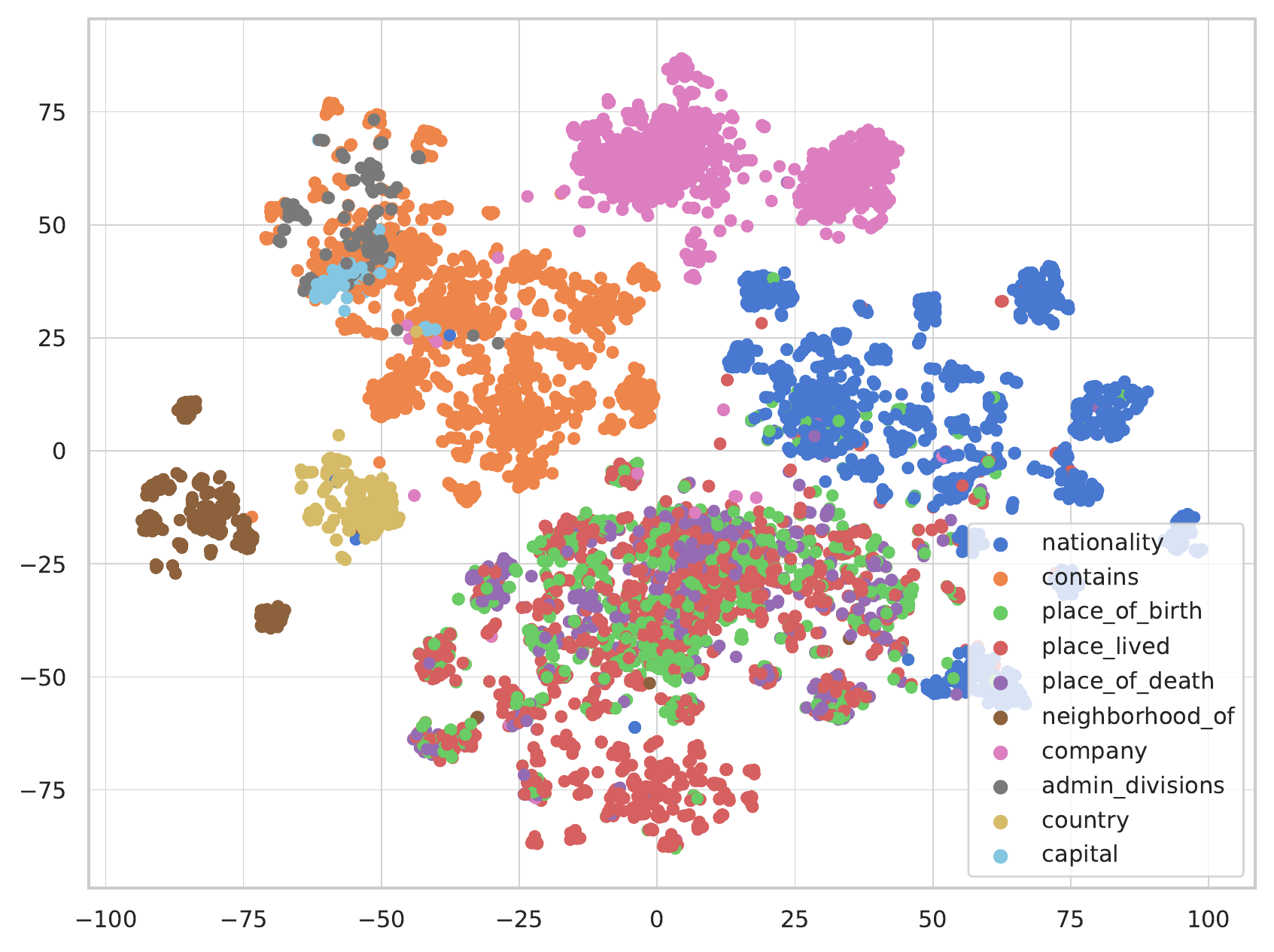}
    \caption{Prior Freebase}
    \label{fig:prior_fb}
    \end{subfigure}%
    \begin{subfigure}[b]{0.25\textwidth}
    \includegraphics[width=\linewidth]{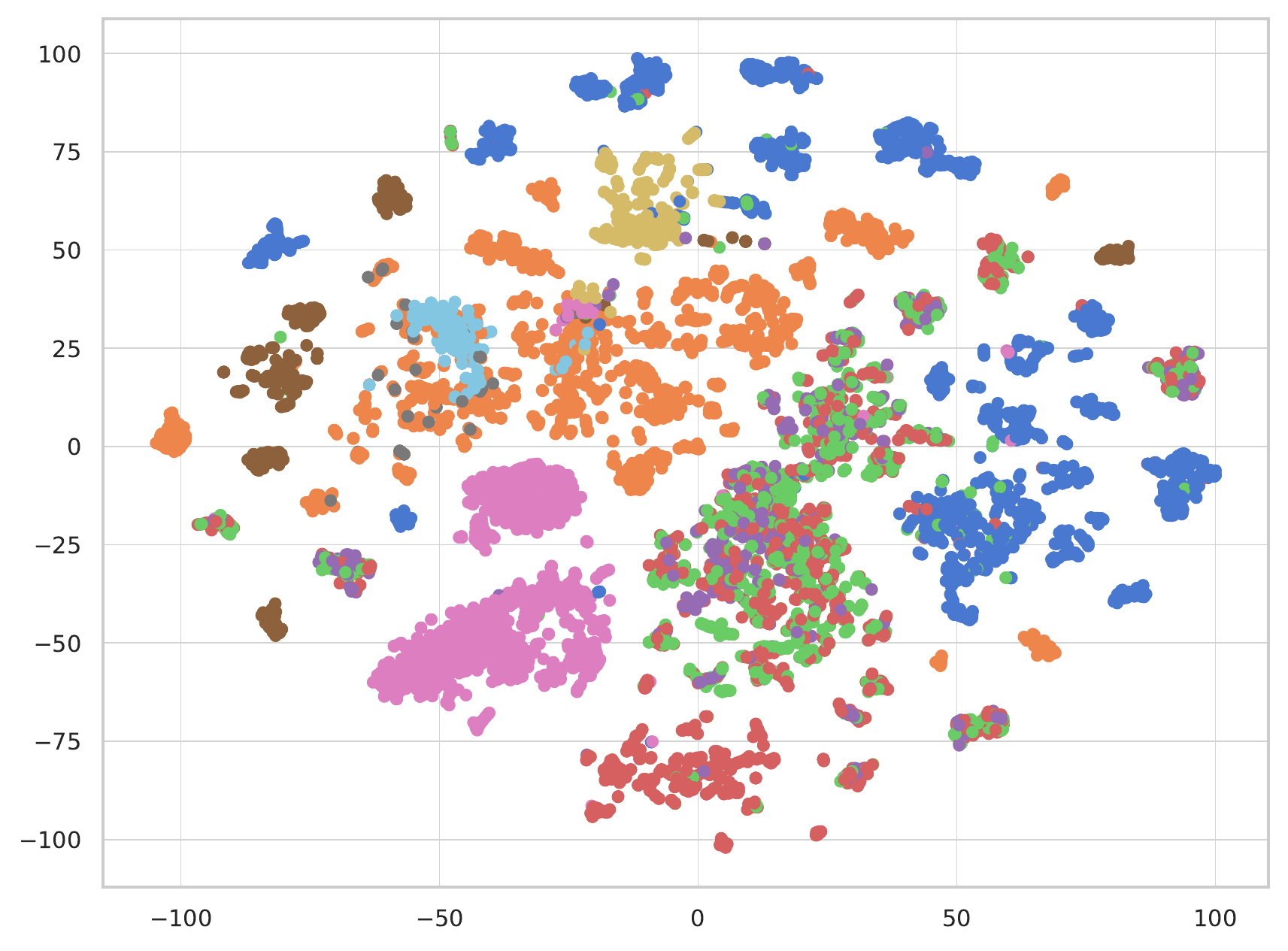}
    \caption{Posterior Freebase}
    \label{fig:post_fb}
    \end{subfigure}%
    \begin{subfigure}[b]{0.25\textwidth}
    \includegraphics[width=\linewidth]{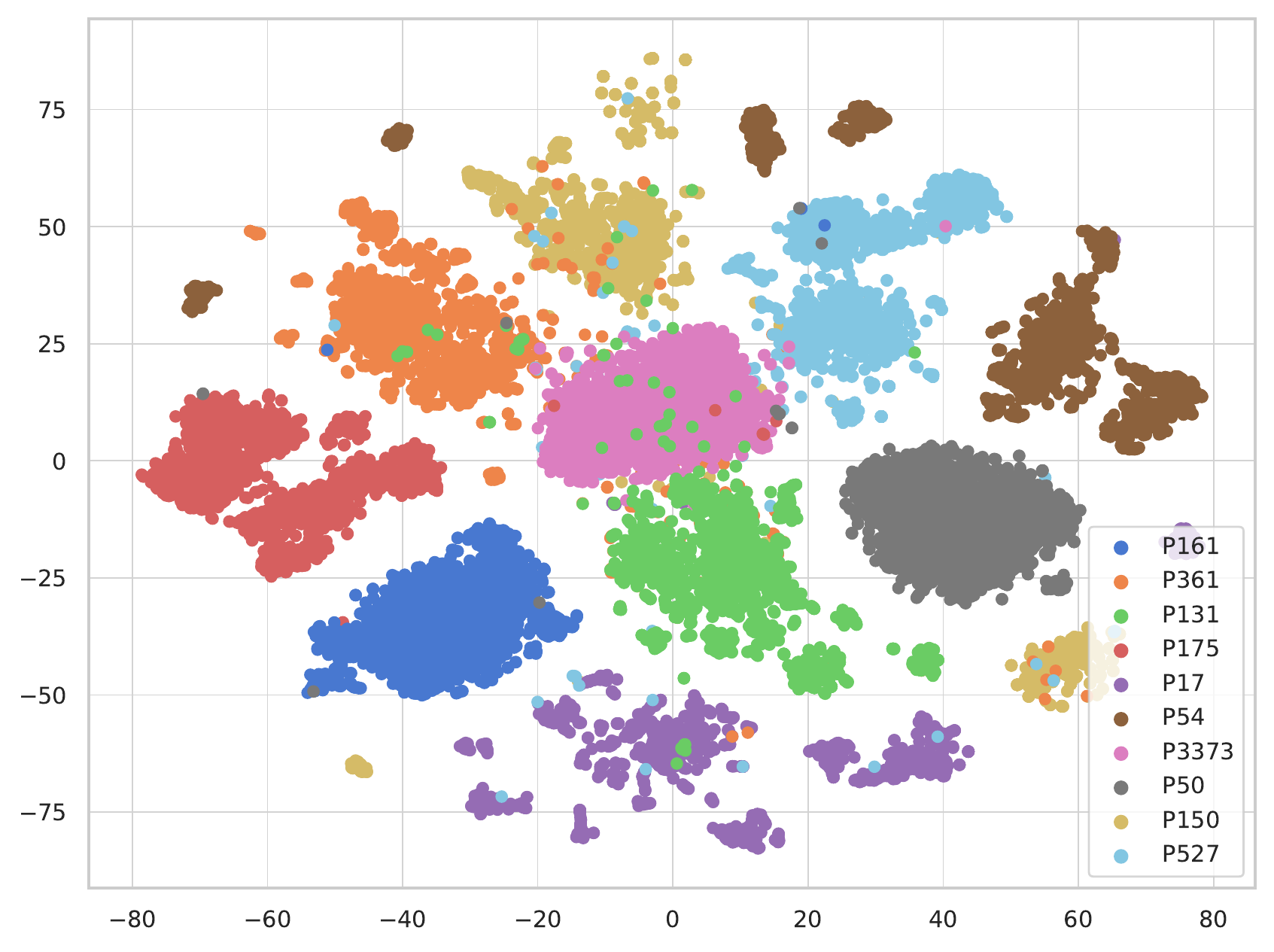}
    \caption{Prior Wikidata}
    \label{fig:prior_wiki}
    \end{subfigure}%
    \begin{subfigure}[b]{0.25\textwidth}
    \includegraphics[width=\linewidth]{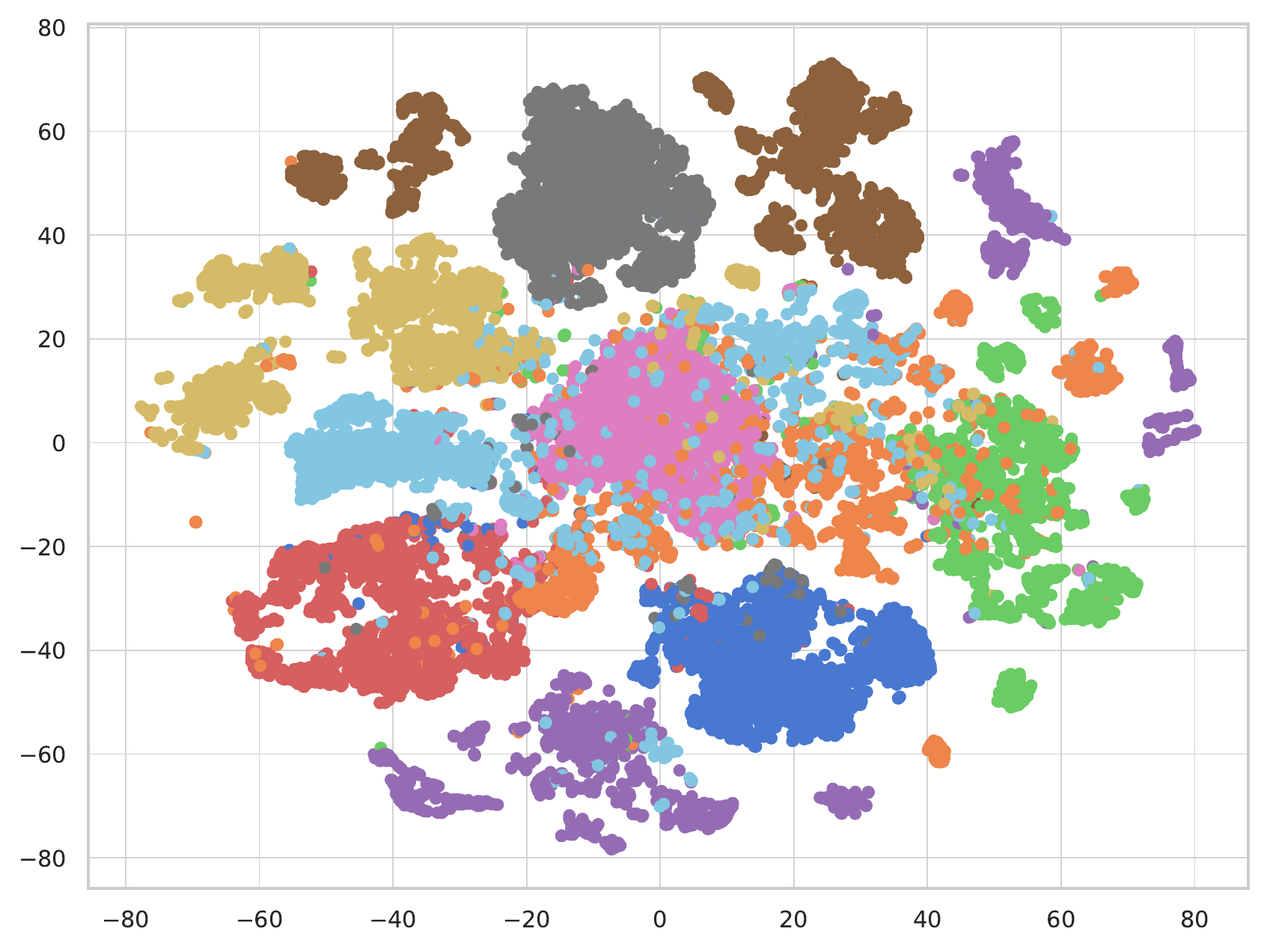}
    \caption{Posterior Wikidata}
    \label{fig:post_wiki}
    \end{subfigure}
    \caption{T-SNE plots of: (a), (c) pair representations obtained from a TransE model (priors) on a subset of Freebase and Wikidata for the 10 most frequent classes in each dataset, (b), (d) the latent codes ($\mu$) for sentences of each training set, when using KB priors.}
\end{figure*}

We then check whether the latent space has indeed learned some information about the KB triples, by visualising the t-SNE plots of the priors, i.e. the $\bm{\mu}_\text{KB}$ vectors as resulted from training TransE (Equation (\ref{eq:mu_kb})) and the posteriors, i.e. the $\bm{\mu}$ vectors as resulted from the \textsc{vae} encoder (Equation (\ref{eq:mu_sigma})).

Figure \ref{fig:prior_fb} illustrates the space of the priors in Freebase for the most frequent relation categories in the \textsc{nyt10} training set
\footnote{We plot t-SNEs for the training set instead of the validation/test sets because the \textsc{WikiDistant} validation set contains too few pairs belonging to the top-10 categories. \textsc{nyt10} validation set t-SNE can be found in the Appendix \ref{app:more_plots}}.
As it can be observed, the separation is obvious for most categories, with a few overlaps. Relations \textit{place of birth}, \textit{place lived} and \textit{place of death} appear to reside in the same region. This is expected as these relations can be shared by a pair simultaneously. Another overlap is identified for \textit{contains}, \textit{administrative divisions} and \textit{capital}. Again, these are similar relations found between certain entity types (e.g. location, province, city).
Figure \ref{fig:post_fb} shows the t-SNE plot for a collection of latent vectors (random selection of 2 sentences in a positive bag). The space is very similar to that of the KB and the same overlapping regions are clearly observed. 
A difference is that it appears to be less compact, as not all sentences in a bag express the exact same relation. 

Similar observations stand for Wikidata priors, as shown in Figure \ref{fig:prior_wiki}.
By looking at the space of the posteriors, we can see that although for most categories separation is achieved, there are 2 relations that are not so well separated in the posterior space. We find that \textit{has part} (cyan) and \textit{part of} (orange) are opposite relations, that TransE can effectively learn thanks to its properties. However, the model appears to not be able to fully separate the two. These relations are expressed in the same manner, by only changing the order of the arguments. As there is no restriction regarding the argument order in our model directionality can sometimes be an issue.

\begin{table*}[ht!]
    \centering
    \setlength{\tabcolsep}{7pt}
    \scalebox{0.75}{
    \begin{tabular}{lrp{8.9cm}p{7.3cm}}
        \hline
        & \tt{\small INPUT} 
        & she graduated from \textit{\_ college} in \textit{new concord} , ohio 
        & growing up in \textit{bay village} , \textit{ohio} , steinbrenner haunted the county fairs , riding in pony races .  \\ \hline
        
        \multirow{3}{*}{\small{$\mathcal{N}(\bm{0}, \mathbf{I})$}} & \tt{\small MEAN} 
        & he graduated from the university of california and received a master 's degree in education . 
        & he was born in \_ , england , and grew up in the united states  \\
        
        & \tt{\small SAMPLE} 
        & he graduated from the university of california and received a master 's degree in education . 
        & he was born in \# , and then moved to new york   \\ \hline

        \multirow{4}{*}{\small{$\mathcal{N}(\bm{\mu_\textsc{kb}}, \mathbf{I})$}} & \tt{\small MEAN} 
        & the bridegroom , \# , is a professor of the university of california at berkeley , and a professor of english ...
        & the \_ , which is based in new york , and the \_ ... \\
        
        & \tt{\small SAMPLE} 
        & the \_ , a \_ of the university of california , berkeley , and the author of '' the \_ of the world  '' ...
        & the \_ , which is based in new jersey , and the \_ ... \\ \hline
    \end{tabular}
    }
    \caption{Sentence reconstruction examples from the \textsc{nyt10} validation set, using different priors. \_ corresponds to the UNK word and \# indicates a number.}
    \label{tab:reco_fb}
\end{table*}

\begin{table*}[ht!]
    \centering
    \setlength{\tabcolsep}{7pt}
    \scalebox{0.75}{
    \begin{tabular}{lrp{9cm}p{7.2cm}}
        \hline
        & \tt{\small INPUT} 
        & \textit{wayne rooney} plays as a striker for manchester united and the \textit{england national team}
        &  ng 's first role was in the \# \textit{michael hui} comedy film `` \textit{the private eyes} '' . \\ \hline
        
        \multirow{3}{*}{\small{$\mathcal{N}(\bm{0}, \mathbf{I})$}} & \tt{\small MEAN} 
        &  \_ 's first game was the first time in the game against the new york yankees . 
        &  the film was adapted into the \# film ` the \_ ' , directed by \_ . \\
        
        & \tt{\small SAMPLE} 
        &  he made his debut for the club in the \# fa cup final against arsenal at wembley stadium .
        & in \# , he appeared in ` the \_ ' , a \# film adaptation of the same name by \_ .
        \\ \hline
        
        \multirow{3}{*}{\small{$\mathcal{N}(\bm{\mu_\textsc{kb}}, \mathbf{I})$}} & \tt{\small MEAN} 
        & he was a member of the club 's first team , and was a member of the club 's \_ club
        & \_ 's first film was ` the \_ ' , starring \_ and starring \_ .\\
        
        & \tt{\small SAMPLE} 
        & he made his debut in the russian professional football league for fc \_ ... 
        & \_ , who was the first female actress to win the academy award for best actress . \\ \hline
    \end{tabular}
    }
    \caption{Sentence reconstruction examples from the \textsc{WikiDistant} validation set using different priors. \_ corresponds to the UNK word and \# indicates a number.}
    \label{tab:reco_wiki}
\end{table*}

Finally, in order to check how the prior constraints affect sentence reconstruction, we illustrate reconstructions of sentences in the validation set of the \textsc{nyt10} in Table \ref{tab:reco_fb} and \textsc{WikiDistant} in Table \ref{tab:reco_wiki}.
In detail, we give the input sentence to the network and employ greedy decoding using either the mean of the latent code or a random sample.

Manual inspection of reconstruction reveals that KB-priors generate longer sentences than the Normal prior by repeating several words (especially the UNK). 
In fact, \textsc{vae} with KB-priors fails to generate plausible and grammatical examples for \textsc{nyt10}, as shown in Table \ref{tab:reco_fb}. 
Instead, reconstructions for \textsc{WikiDistant} are slightly better, due to the less noisy nature of the dataset.
In both cases, we see that the reconstructions contain words that are useful for the target relation, e.g. words that refer to places such as \textit{new york}, \textit{new jersey} for the relation \textit{contains} between \textit{bay village} and \textit{ohio}, or sport-related terms (football, team, league) for the \textit{statistical leader} relationship between \textit{wayne rooney} and \textit{england national team}.

\section{Related Work}

\textbf{Distantly Supervised RE.} 
Methods developed for DSRE have been around for a long time, building upon the idea of distant supervision~\citep{mintz2009distant} with the widely used \textsc{nyt10} corpus by \citet{riedel2010modeling}. 
Methods investigating this problem can be divided into several categories. Initial approaches were mostly graphical models, adopted to perform multi-instance learning~\citep{riedel2010modeling}, sentential evaluation~\citep{hoffmann2011knowledge,bai2019structured} or multi-instance learning and multi-label classification~\citep{surdeanu2012multi}. 
Subsequent approaches utilised neural models, with the approach of \citet{zeng2015distant} introducing Piecewise Convolutional Neural Networks (PCNN) into the task.  
Later approaches focused on noise reduction via selection of informative instances using either soft constraints, i.e., attention mechanisms~\citep{lin2016neural,ye2019intra-inter,yuan2019cross}, or hard constraints by explicitly selecting non-noisy instances with reinforcement~\citep{feng2018reinforcement,qin2018robust,qin2018dsgan,wu2019improving,yang2019exploiting} and curriculum learning~\citep{du2019curriculum}.
Noise at the word level was addressed in \citet{liu2018neural} via sub-tree parsing on sentences.
Adversarial training has been shown to improve DSRE in \citet{wu2017adversarial}, while additional unlabelled examples were exploited to assist classification with Generative Adversarial Networks (GAN)~\citep{goodfellow2014generative} in \citet{li2019gan}.
Recent methods use additional information from external resources such as entity types and relations~\citep{vashishth2018reside}, entity descriptors~\citep{ji2017descriptors,she2018descriptions,hu2019improving} or Knowledge Bases~\citep{weston2013connecting,xu2019hrere,li2020self}. \\

\noindent \textbf{Sequence-to-Sequence Methods.} 
Autoencoders and variational autoencoders have been investigated lately for relation extraction, primarily for detection of relations between entity mentions in sentences. \citet{marcheggiani2016discrete} proposed discrete-state \textsc{vae}s for link prediction, reconstructing one of the two entities of a pair at a time.
\citet{ma2019mcvae} investigated conditional \textsc{vae}s for sentence-level relation extraction, showing that they can generate relation-specific sentences. Our overall approach shares similarities with this work since we also use \textsc{vae}s for RE, though in a bag rather than a sentence-level setting.
\textsc{vae}s have also been investigated for RE in the biomedical domain~\citep{zhang2019exploring}, where additional non-labelled examples were incorporated to assist classification. 
This work also has commonalities with our work but the major difference is that the former uses two different encoders while we use only one, shared among bag classification and bag reconstruction.
Other \textsc{seq2seq} methods treat RE as a sequence generation task. Encoder-decoder networks were proposed for joint extraction of entities and relations~\citep{trisedya2019neural,nayak2020effective}, generation of triples from sequences~\citep{liu2018seq2rdf} or generation of sequences from triples~\citep{trisedya2018gtr,zhu2019triple}. \\

\noindent \textbf{\textsc{vae} Priors.}
Different types of prior distributions have been proposed for \textsc{vae}s, such as the VampPrior~\citep{tomczak2018vae}, Gaussian mixture priors~\citep{dilokthanakul2016deep}, Learned Accept/Reject Sampling (LARs) priors~\citep{bauer2019resampled}, non-parametric priors~\citep{goyal2017nonparametric} and others.
User-specific priors have been used in collaborative filtering for item recommendation~\citep{karamanolakis2018item}, while topic-guided priors were employed for generation of topic-specific sentences~\citep{wang2019topic}.
In our approach we investigate how to incorporate KB-oriented Gaussian priors in DSRE using a link prediction model to parameterise their mean vector.

\section{Conclusions} 
We proposed a probabilistic approach for distantly supervised relation extraction, which incorporates context agnostic knowledge base triples information as latent signals into context aware bag-level entity pairs.
Our method is based on a variational autoencoder that is trained jointly with a relation classifier. KB information via a link prediction model is used in the form of prior distributions on the \textsc{vae} for each pair. The proposed approach brings close sentences that contain the same KB pairs and it does not require any external information during inference time.

Experimental results suggest that jointly reconstructing sentences with relation classification is helpful for distantly supervised RE and KB priors further boost performance. Analysis of the generated latent representations showed that we can indeed manipulate the space of sentences to match the space of KB triples, while reconstruction is enforced to keep topic-related terms.   

Future work will target experimentation with different link prediction models and handling of non-informative sentences. 
Finally, incorporating large pre-trained language models (LMs) into \textsc{VAE}s is a recent and promising study~\citep{li-etal-2020-optimus} which can be combined with KBs as injecting such information into LMs has been shown to further improve their performance~\citep{peters-etal-2019-knowledge}.

\section*{Acknowledgements}
This research was supported by BBSRC Japan Partnering Award [Grant ID: BB/P025684/1] and based on results obtained from a project, JPNP20006, commissioned by the New Energy and Industrial Technology Development Organization (NEDO). The authors would like to thank the anonymous reviewers for their instructive comments.

\bibliography{anthology,custom}
\bibliographystyle{acl_natbib}

\appendix

\section{Appendix}

\subsection{The \textsc{nyt10} Dataset}
\label{app:info_nyt}
As described in \citet{bai2019structured}, the \textsc{nyt10} dataset has been released in several versions. The original one, follows the setting of \citet{riedel2010modeling}, where two sets of data were created. Later versions~\citep{lin2016neural} merged the two sets in order to construct a larger dataset. This merging resulted into $570,300$ instances for training. However, in this version of the data exist overlaps in pairs between the training and the test set.
The amount of overlaps is significant and accounts for $47,477$ instances, which is approximately $27.5\%$ of the testing instances.
The version was corrected later on but there still remain methods that use the non-filtered data.
Recently, \citet{han2019opennre} released a finalised version removing the overlaps, resulting in $522,611$ total training instances.
In our experiments we evaluate the proposed model on both versions.

It is also important to note that \textsc{nyt10} has been used by the community in two settings: bag-level and sentence-level.
In the bag-level setting, a pair's relation is defined based on a bag of sentences that contain the pair. 
On the contrary, in the sentence-level setting a pair's relation is predicted for each sentence. Training data are obtained using distant supervision, while test data are manually annotated~\citep{hoffmann2011knowledge}.

\subsection{Data Pre-processing Details}

We found that the dataset includes several duplicate instances, i.e. the exact same sentence with the exact same pair. We remove such cases from our training data since they can bias the training process. However, they are preserved on the validation and test sets for a fair comparison with other methods.
We convert the dataset to lowercase and replace all digits with the hash character (\#). We randomly select $10\%$ of the training bags as our validation set. \\

\begin{table}[h!]
    \centering
    \scalebox{0.85}{
    \begin{tabular}{llrrr}
    \toprule
        & & Train & Validation & Test \\ \midrule
        \parbox[t]{2mm}{\multirow{4}{*}{\rotatebox[origin=c]{90}{Processed}}}
        & Instances      & 400,100 & 53,319 & 172,448 \\
        & Bags           & 248,352 & 28,108 & 96,678 \\ 
        & Facts          & 16,338  & 1,823  & 1,950 \\
        & Negatives      & 233,092 & 26,301 & 94,917 \\ \midrule
        & Instances      & 469,290 & 53,321 & - \\
        & Bags           & 252,044  & 28,109  & - \\
        & Duplicates     & 62,327  & -  & - \\
        & Outliers       & 5,570   & -  & - \\
    \bottomrule
    \end{tabular}
    }
    \caption{Statistics of the \textsc{nyt10} (520K version) dataset.}
    \label{tab:nyt_stats}
\end{table}

\begin{table}[h!]
    \centering
    \scalebox{0.85}{
    \begin{tabular}{llrrr}
    \toprule
        & & Train & Validation & Test \\ \midrule
        \parbox[t]{2mm}{\multirow{4}{*}{\rotatebox[origin=c]{90}{Processed}}}
        & Instances      & 434,453 & 62,333 &  172,448 \\
        & Bags           & 258,843 & 29,303 &  96,678 \\ 
        & Facts          & 17,387 & 1,942 &  1,950\\
        & Negatives      & 242,644 & 27,374 & 94,917\\ \midrule
        & Instances      & 507,755 & -  & - \\
        & Bags           & 262,649 & -  & - \\
        & Duplicates     & 66,130 & -   & - \\
        & Outliers       & 5,856 & -    & - \\
    \bottomrule
    \end{tabular}
    }
    \caption{Statistics of the \textsc{nyt10} (570K version) dataset.}
    \label{tab:nyt570_stats}
\end{table}

\begin{table}[h!]
    \centering
    \scalebox{0.85}{
    \begin{tabular}{llrrr}
    \toprule
        & & Train & Validation & Test \\ \midrule
        \parbox[t]{2mm}{\multirow{4}{*}{\rotatebox[origin=c]{90}{Processed}}}
        & Instances       & 1,000,765 &  29,145 & 28,897 \\
        & Bags            & 572,215   &  14,748 & 15,509 \\ 
        & Facts           & 201,356   &  4,333  & 4,333 \\
        & Negatives       & 370,859   &  10,415 & 11,176 \\ \midrule
        & Instances       & 1,050,246 &  - &  - \\
        & Bags            & 575,620   & - & - \\
        & Duplicates      & 43,978    &  - &  - \\
        & Outliers        & 5,503     &  - &  - \\
    \bottomrule
    \end{tabular}
    }
    \caption{Statistics of the \textsc{WikiDistant} dataset.}
    \label{tab:wiki_stats}
\end{table}

\noindent \textbf{Sentence Length Filtering.} 
We restrict the length of a sentence to 50 words for the \textsc{nyt10} dataset and to 30 for the \textsc{WikiDistant} dataset.
If at least one of the arguments of a pair is located in a span after the maximum sentence length, then the sentence is resized to contain the words from the first argument until the second. We also add a maximum number of 5 words to the left and 5 words to the right if the total length allows. 
If the length of the resized sentence is still larger than the maximum sentence length, the sentence is removed from the training set.
The reason for this choice is that we want to construct contextualised argument representations. Without the arguments inside the sentence, such representations cannot be formed. We call such removed sentences \textit{outliers}.
Outliers are not removed for the validation and test sets. 
Relevant statistics are shown in Tables \ref{tab:nyt_stats}, \ref{tab:nyt570_stats} and \ref{tab:wiki_stats}. \\

\noindent \textbf{Vocabulary construction.} 
In order to construct the word vocabulary, we use the unique sentences contained in the training set, as resulted from the removal of duplicate instances and the sentence length filtering. 
Since each sentence in the dataset can contain multiple pairs, it is repeated for each pair. 
Using non-unique sentences can lead to counting larger frequencies for certain words and producing a misleading vocabulary. 
We restrict the vocabulary to contain the 40K most frequent words for \textsc{nyt10}, with a coverage of $97.78\%$ in the training set and to 50K for \textsc{WikiDistant} with a coverage of $96\%$.
Other words are replaced with the UNK token.

\subsection{Hyper-parameter Settings}

\textbf{DSRE Models.}
Table \ref{tab:nyt_hyper} shows the parameters used for training the model on the \textsc{nyt10} and \textsc{WikiDistant} dataset. 
In the \textsc{vae} setting Adaptive Softmax \citep{grave2017adaptive} was incorporated instead of regular Softmax for faster training. We used three clusters by splitting the vocabulary in
$\floor{\frac{|V|}{15}}$ and $\floor{\frac{3|V|}{15}}$ words. \\

\begin{table}[h!]
    \centering
    \scalebox{0.85}{
    \begin{tabular}{lrr}
    \toprule
        Parameter & \textsc{nyt} & \textsc{wiki}  \\ \midrule
        Batch size & 128 & 128 \\
        Max bag size & 500 & 500 \\
        Learning rate & 0.001 & 0.001 \\
        Weight decay & $10^{-6}$ & $10^{-6}$ \\
        Gradient clipping & 10 & 5 \\
        Optimiser & Adam & Adam \\
        Early stopping patience & 5 & 5 \\
        Task loss weight $\lambda$       & 0.8, 0.9 & 0.9 \\ \midrule
        Word embedding $\mathbf{E}^{(w)}$ dim. & 50 & 50 \\
        Relation embedding $\mathbf{E}^{(r)}$ dim. & 64 & 128 \\
        Position embedding $\mathbf{E}^{(p)}$ dim. & 8 & 8 \\
        Latent code $z$ dim.        & 64 & 64 \\
        Teacher force &  0.3  & 0.3 \\
        Encoder dim. & 256 & 256 \\
        Encoder layers & 1 & 1 \\
        Decoder dim. & 256 & 256 \\
        Decoder layers & 1 & 1 \\
        Input dropout & 0.3 & 0.3 \\
        Word dropout & 0.3 & 0.1 \\
    \bottomrule
    \end{tabular}
    }
    \caption{Models hyper-parameters for each dataset.}
    \label{tab:nyt_hyper}
\end{table}

\noindent \textbf{Knowledge Base Embeddings.}
In order to train KB entity embeddings we used the DGL-KE toolkit~\citep{DGL-KE}. We use the same set of hyper-parameters for both Freebase and Wikidata as shown in Table \ref{tab:kb_hyper}. For Freebase we select $5,000$ triples as the validation set, while for Wikidata we use the validation set provided in the transductive setting ($5,136$ triples).

\begin{table}[h!]
    \centering
    \scalebox{0.8}{
    \begin{tabular}{lr}
    \toprule
        Parameter & Value \\ \midrule
        Model & TransE\_l2 \\
        Emb. size & 64 \\
        Max train step & 500,000 \\
        Batch size & 1024 \\
        Negative sample size & 256 \\
        Learning rate & 0.1 \\
        Gamma & 10.0 \\
        Negative adversarial sampling & True \\
        Adversarial temperature & 1.0 \\
        Regularisation coefficient & $10^{-7}$ \\
        Regularisation norm & 3 \\
    \bottomrule
    \end{tabular}
    }
    \caption{Knowledge Base Embeddings hyper-parameters.}
    \label{tab:kb_hyper}
\end{table}

\subsection{\textsc{WikiDistant} Relation Categories}
Since \textsc{WikiDistant} contains 454 relations, their labels are used directly from the WikiData properties\footnote{\url{https://www.wikidata.org/wiki/Wikidata:List_of_properties}}.
Here, we add explanations about the top 10 most frequent categories used in Figures \ref{fig:prior_wiki}, \ref{fig:post_wiki}.

\begin{table}[h!]
    \centering
    \scalebox{0.8}{
    \begin{tabular}{lp{4.5cm}}
    \toprule
        P17 & country  \\
        P3373 & sibling \\
        P131 & located in the administrative territorial entity \\
        P54  & member sports team \\
        P175 & performer \\
        P161 & cast member \\
        P361 & part of \\
        P50 & author \\
        P150 & contains administrative territorial entity \\
        P527 & has part \\
    \bottomrule
    \end{tabular}
    }
    \caption{Explanations of the top 10 most frequent \textsc{WikiDistant} relation categories.}
    \label{tab:my_label}
\end{table}

\begin{figure*}[t!]
    \centering
    \begin{minipage}[c]{0.45\textwidth}
    \includegraphics[width=0.85\textwidth]{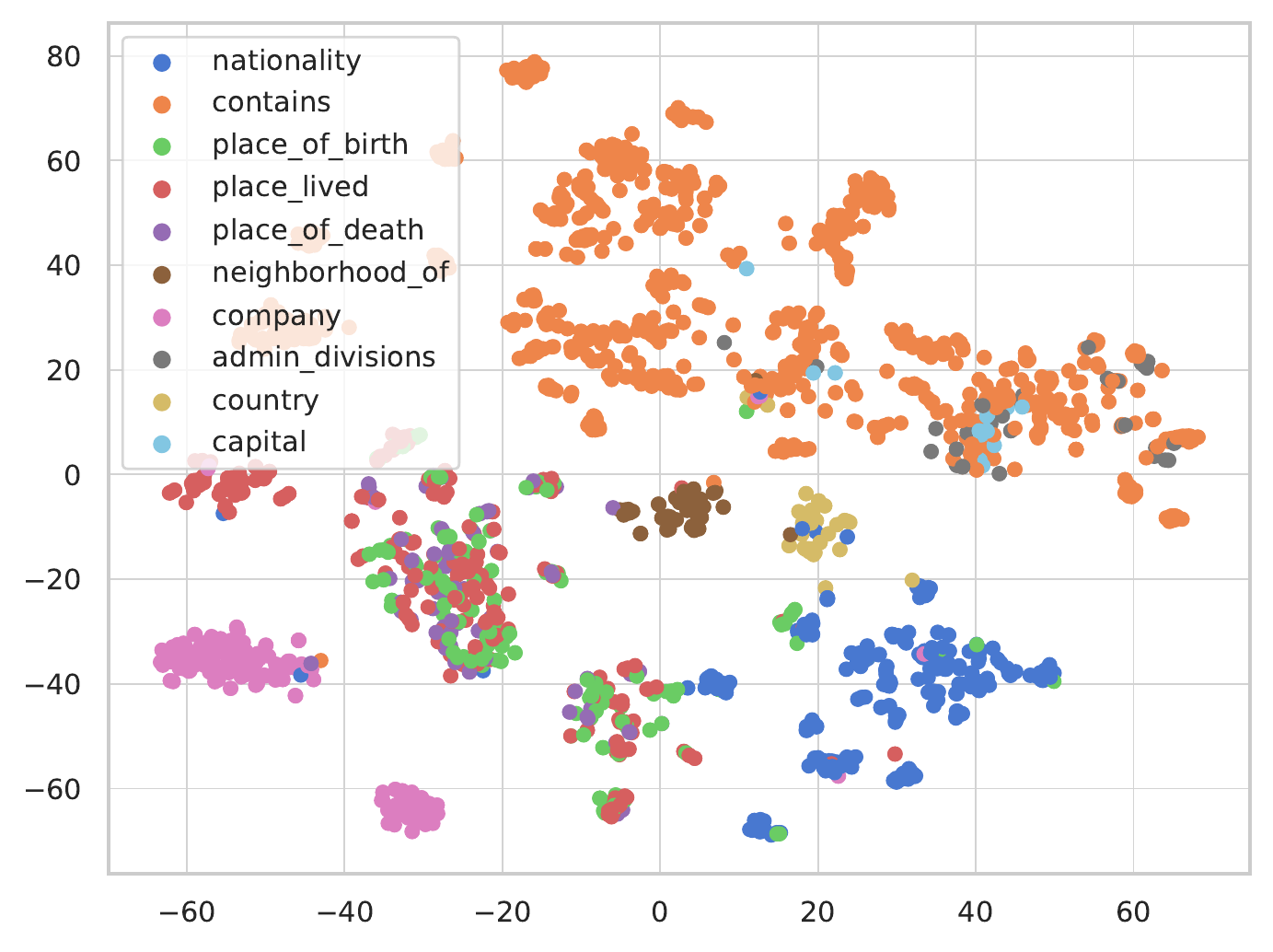}
    \caption{t-SNE plot of the latent vector ($\mu$) for the \textsc{nyt10} (520K) validation set, when using KB priors during training.}
    \label{fig:tsne_nyt10_val}
    \end{minipage}\hspace{0.05\linewidth}
    \begin{minipage}[c]{0.45\textwidth}
    \includegraphics[width=0.85\textwidth]{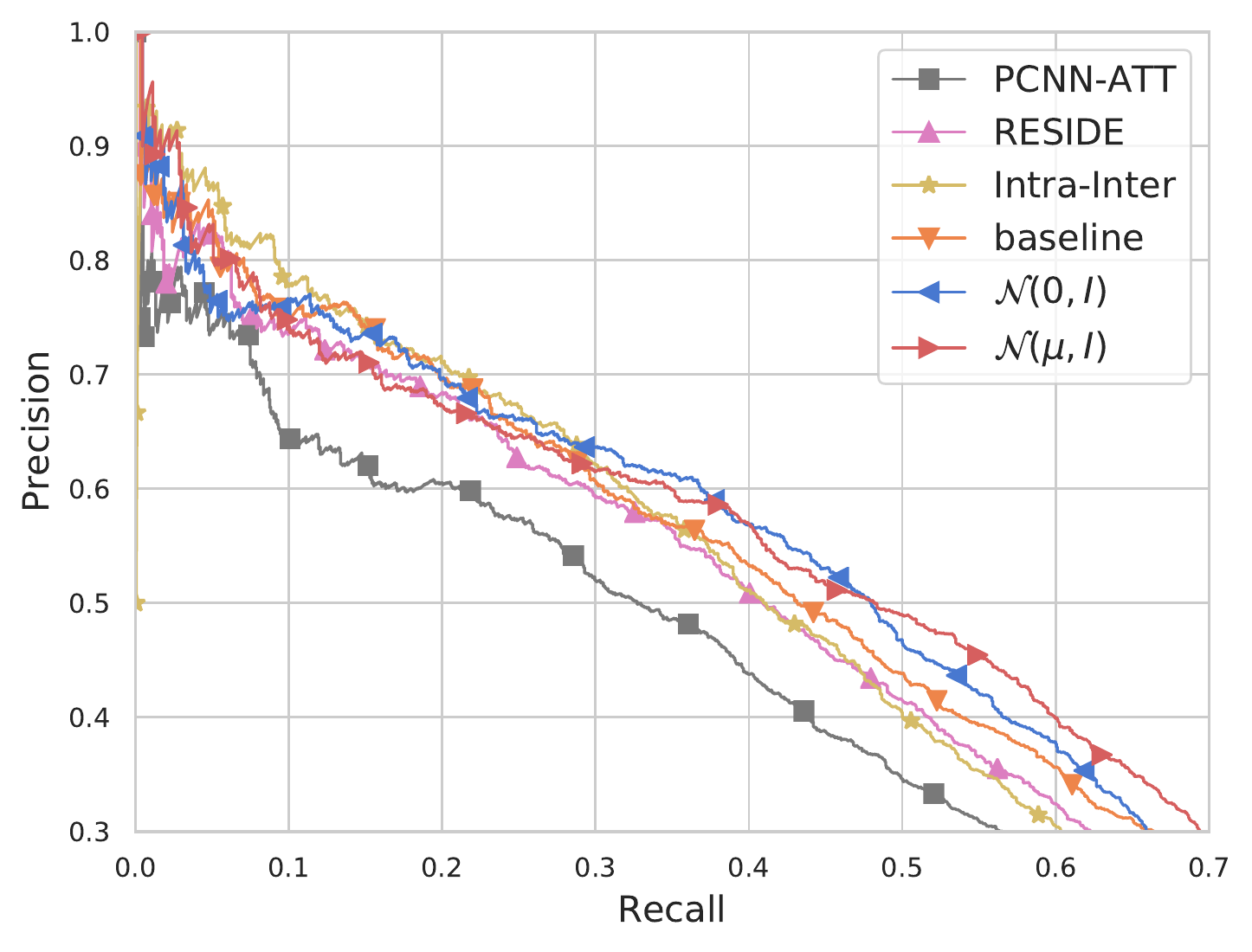}
    \caption{Precision-Recall curves for the \textsc{nyt10} (570K) test set.}
    \label{fig:pr_curves_nyt_570}
    \end{minipage}
\end{figure*}

\subsection{Additional Plots}
\label{app:more_plots}

Figure \ref{fig:tsne_nyt10_val} illustrates the t-SNE plot of the latent space for the \textsc{nyt10} validation set. We observe similar clusters to that of the KB (Figure \ref{fig:prior_fb}).

Figure \ref{fig:pr_curves_nyt_570} illustrates the PR-curves for the non-filtered version of the \textsc{nyt10} dataset (570K). Here, KB-priors perform comparably with Normal prior but mostly improve the tail of the distribution (after 50\% of the recall range). We could not obtain the PR curve for the \textsc{jointnre} method, thus it is not present in the figure.

\end{document}